\documentclass[letterpaper, 10 pt, conference]{ieeeconf}  

\IEEEoverridecommandlockouts                              
\usepackage{graphics} 
\usepackage{epsfig} 
\usepackage{mathptmx} 
\usepackage{times} 
\usepackage{amsmath} 
\usepackage{amssymb}  
\usepackage{stfloats}
\usepackage{color}
\usepackage{soul}
\usepackage{siunitx}
\usepackage{multicol}
\usepackage{float}
\usepackage{url}

\title{Design, Modeling, and Control of a Low-Cost and Rapid Response Soft-Growing Manipulator for Orchard Operations}

\author{Ryan Dorosh, Justin Allen, Zixuan He, Christopher Ninatanta, Jack Coleman, Jack Spieker, Ethan Tuck,\\ Jordan Kurtz, Qin Zhang, Matthew D. Whiting, Jiecai Luo, Manoj Karkee, and Ming Luo, \emph{Member, IEEE}
\thanks{This work was supported by National Institute of Food and Agriculture 1029004 and Washington Tree Fruit Research Commission. Corresponding author: Ming Luo. (email: ming.luo@wsu.edu)}
\thanks{R.~Dorosh, J.~Allen, C.~Ninatanta, J.~Coleman, J.~Spieker, E.~Tuck, J.~Kurtz, and M.~Luo are with School of Mechanical and Materials Engineering, Washington State University, WA 99163, USA)\newline \indent
Z.~He, Q.~Zhang, and M.~Karkee are with the Department of Biological Systems Engineering, Washington State University, WA 99350, USA.
\newline \indent
M.~Whiting is with the Department of Horticulture, Washington State University, WA 99350, USA.
\newline \indent
J.~Luo is with Electrical Engineering Department, Southern University and A\&M College Baton Rouge, LA 70807, USA.}
}

\begin{document}
\maketitle
\thispagestyle{empty}
\pagestyle{empty}

\begin{abstract}
Tree fruit growers around the world are facing labor shortages for critical operations, including harvest and pruning. There is a great interest in developing robotic solutions for these labor-intensive tasks, but current efforts have been prohibitively costly, slow, or require a reconfiguration of the orchard in order to function. In this paper, we introduce an alternative approach to robotics using a novel and low-cost soft-growing robotic platform. Our platform features the ability to extend up to 1.2 m linearly at a maximum speed of 0.27 m/s. The soft-growing robotic arm can operate with a terminal payload of up to 1.4 kg (4.4 N), more than sufficient for carrying an apple. This platform decouples linear and steering motions to simplify path planning and the controller design for targeting. We anticipate our platform being relatively simple to maintain compared to rigid robotic arms. Herein we also describe and experimentally verify the platform’s kinematic model, including the prediction of the relationship between the steering angle and the angular positions of the three steering motors. Information from the model enables the position controller to guide the end effector to the targeted positions faster and with higher stability than without this information. Overall, our research show promise for using soft-growing robotic platforms in orchard operations.

\end{abstract}

\section{Introduction}
\label{sec:Introduction}

The tree fruit industry is a significant sector of the US agricultural industry, representing approximately 10\% of all crop production \cite{calvin2010us}. Tree fruit growers in Washington State lead the nation in the production of apples and sweet cherries, contributing $>\$4$ billion and about $\$1$ billion respectively to the US GDP in 2021 through direct economic activities \cite{nassusda}. Sustained productivity in these tree fruit orchards requires time-sensitive and labor-intensive operations such as pruning, thinning, and harvesting. Farmers depend upon having temporary skilled workers available to carry out these operations. However, due to increasing costs and labor shortages, the industry is becoming increasingly harder to sustain. For example, the labor used in apple harvesting costs the industry between 28.5\% and 34.8\% of total labor cost depending on the apple variety \cite{calvin2022supplement}. Farmers have even had to abandon fruit in their orchards from not being able to secure sufficient labor \cite{martin2007farm}. New and innovative approaches are needed to reduce the extraordinary dependency of tree fruit farmers on labor.
\begin{figure}
	\centering
	\includegraphics[width=1\linewidth]{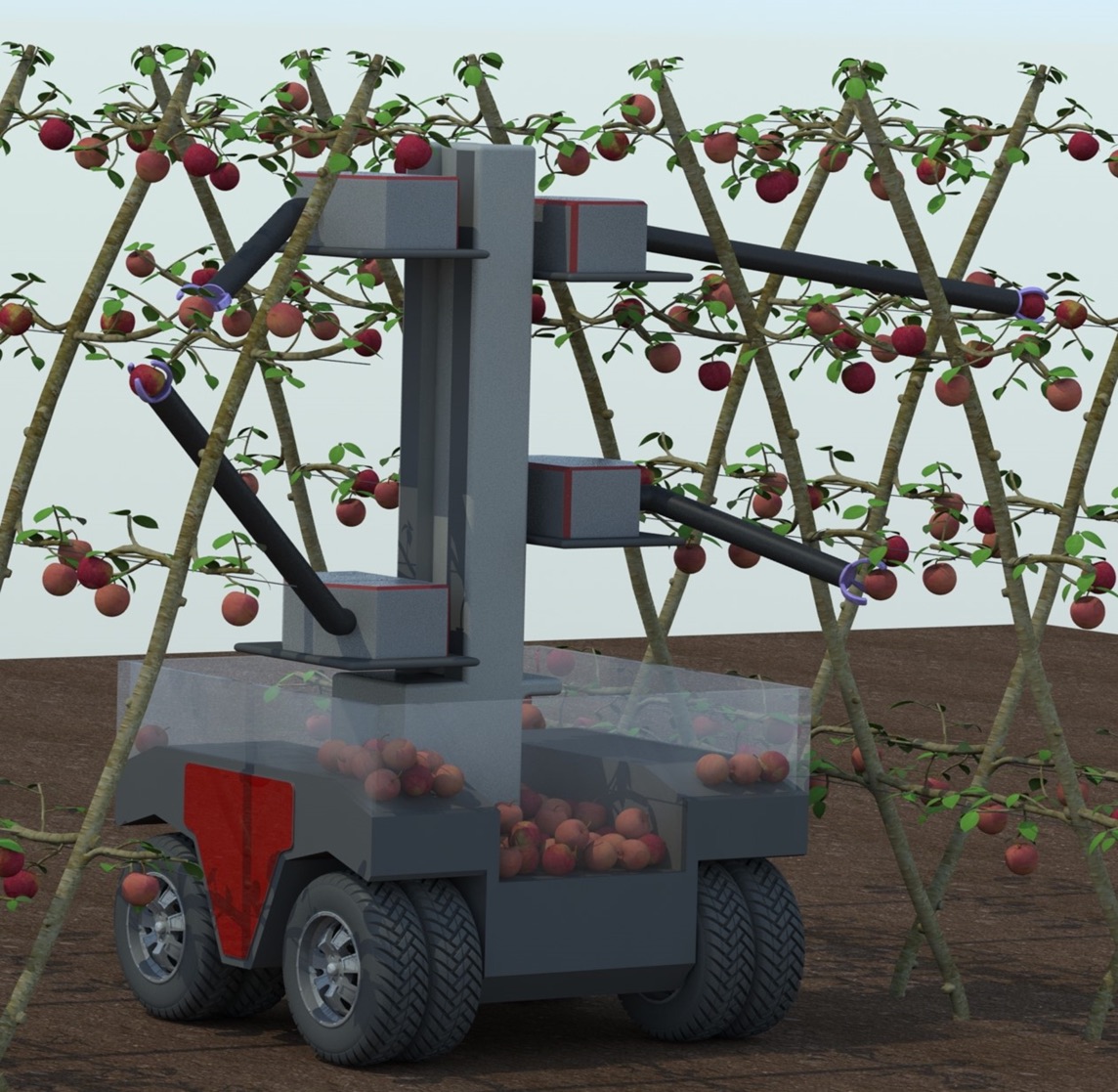}
	\caption{Conceptual depiction of a multi-soft growing manipulator arm system to achieve orchard operations.}
	\label{concept}
	\vspace{-0.2 in}
\end{figure}

Recent efforts have been spent toward the development of robotic solutions to automate critical orchard operations, especially concerning harvesting. Many of these efforts have used widely available rigid-body robots such as the 6-degree of freedom UR5 robotic arm \cite{kang2020real}. These kinds of robotic systems are expensive and unnecessarily complicated for carrying out key operations in orchards. Other efforts, such as the work done by FFRobotics \cite{ffrobotics} and Advanced Farm \cite{advancedfarm}, have utilized linear actuators to minimize the complexity while achieving a long workspace range. The FFRobotics robotic apple harvesting prototype uses numerous large linear actuators that adjust their x and y position along rails before extending to grasp the fruit. This process takes time due to the mass of the arm and rail being moved, and in this configuration, the angle of the arms (i.e., the angle of approach to the target) is fixed. The robotic system by Advanced Farm has more miniature linear actuators that can adjust their angle, yet it still has to adjust the actuators' x and y positions before rotating the entire rail to adjust the angle. Therefore, while these two recent efforts present an improvement on previous work, these methods still require significant time to pick apples due to how they are designed.

In this paper, we describe a novel soft-growing manipulator arm as a potential robotic solution (Fig.~\ref{concept}) to address the labor shortage. The robot's everting mechanism was first described by Okamura's group \cite{hawkes2017soft}. Our previous work with soft robotics demonstrated that human gestures can operate a similar soft-growing manipulator via a teleoperation interface to achieve picking-placing tasks \cite{9197094}. Compared with currently proposed robotic solutions for orchard operations, our robot has the following advantages: 1. Faster linear extension speed at 0.27~m/s at 8~psi (55.1~kpa), meanings that the manipulator can steer at its base to reach its goal within 2s per picking operation in the workspace. 2. Easier to plan and control. The manipulator has two controllers: A length controller actuated by one DC motor and a steering controller actuated by three DC motors. These two controllers are decoupled and act independently. 3. Like other soft robots \cite{zongxing2020research,shiva2016tendon}, the soft material of the arm safely interacts with humans and fruit trees, reducing the planning burden and allowing for more human-robot collaboration to improve working efficiency. 4. Easier to maintain and fix, the maximum input pressure the fabric material can take safely is 18~psi, which is well above our operating pressure of 9~psi, and the fabric can be easily replaced or patched. 5. Low cost at approximately \$4,230, which is about one-eighth the cost of typical rigid commercial manipulators. The cost is an important factor to consider because of the short apple-picking season and large orchard sizes.

In this paper, we demonstrate the design, fabrication, and functionality of a soft-growing manipulator arm for robotic orchard operations in the future. This paper is organized as follows: Section~\ref{sec:Introduction} provides the necessary background and motivation. Section~\ref{sec:design and fabrication} discusses the design, fabrication, and specifications of the soft-growing manipulator arm for orchard operations. Section~\ref{sec:Modeling and control} displays the static modeling and controller design of the manipulator. Section~\ref{sec:Experiment} shows the system diagram, model verification by experimental results, and implementation and analysis of the end effector's position controller. Finally, Section~\ref{sec:conclusion} analyzes the issues present in the current system and outlines paths for future development and research.

In general, the novelty of this work is the utilization of the idea of a soft-growing robot to investigate a low-cost, rapid response, and robust robotic solution for orchard operations. The long-term goal of our team's research is to achieve full autonomy instead of semi-autonomy \cite{9197094} for orchard operations.

\section{Design, Fabrication, and Specifications} 
\label{sec:design and fabrication}

\subsection{Design Overview}

\begin{figure}
	\centering
        \vspace{0 in}
	\includegraphics[width=1\linewidth,]{ 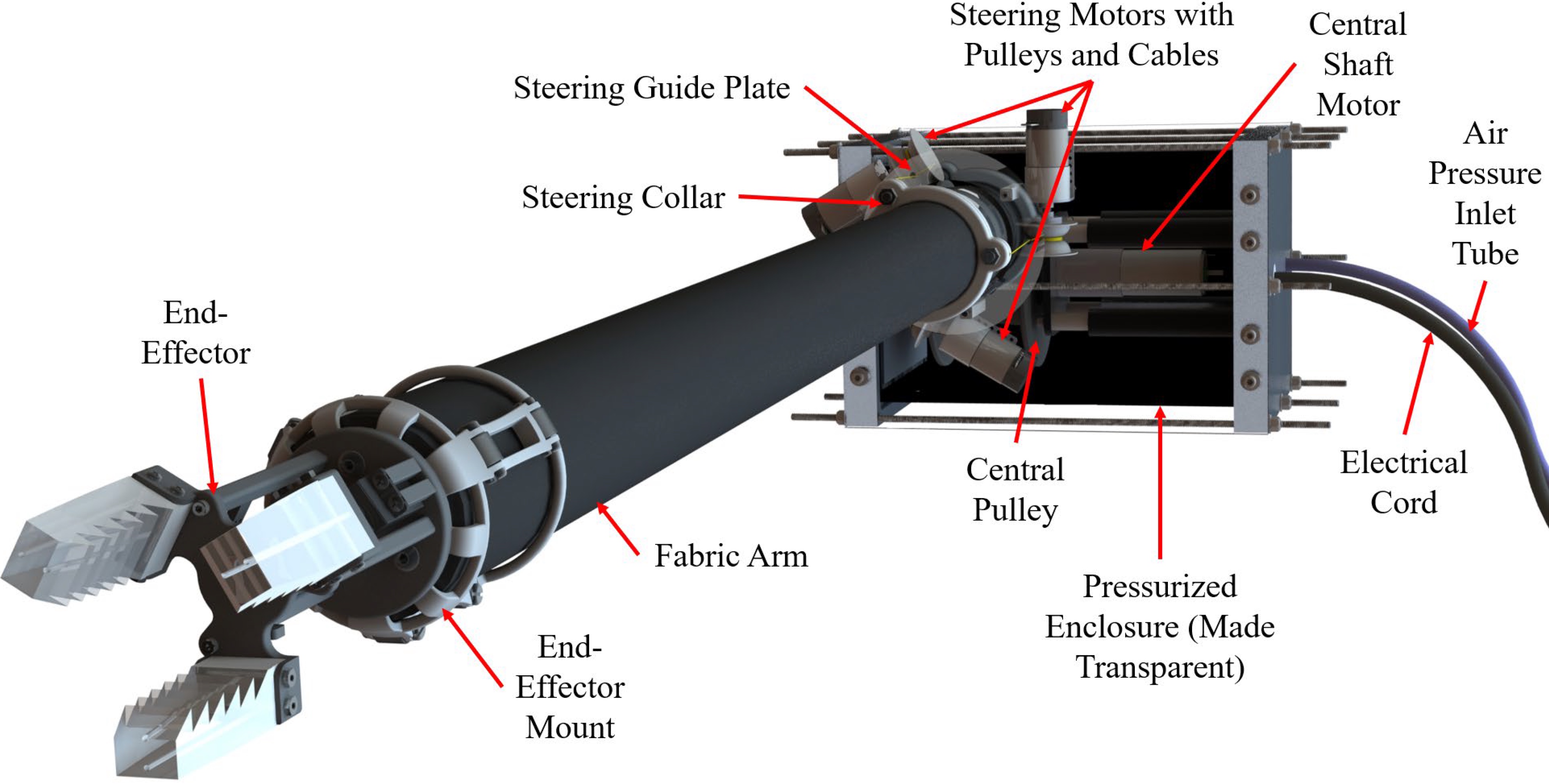}
	\caption{CAD model of the Soft-Manipulator Arm with transparent parts for clarity.}
	\label{cadmodel}
	\vspace{-0.1 in}
\end{figure}

\begin{figure}
	\centering
        \vspace{-0.2 in}
	\includegraphics[width=1\linewidth,]{ 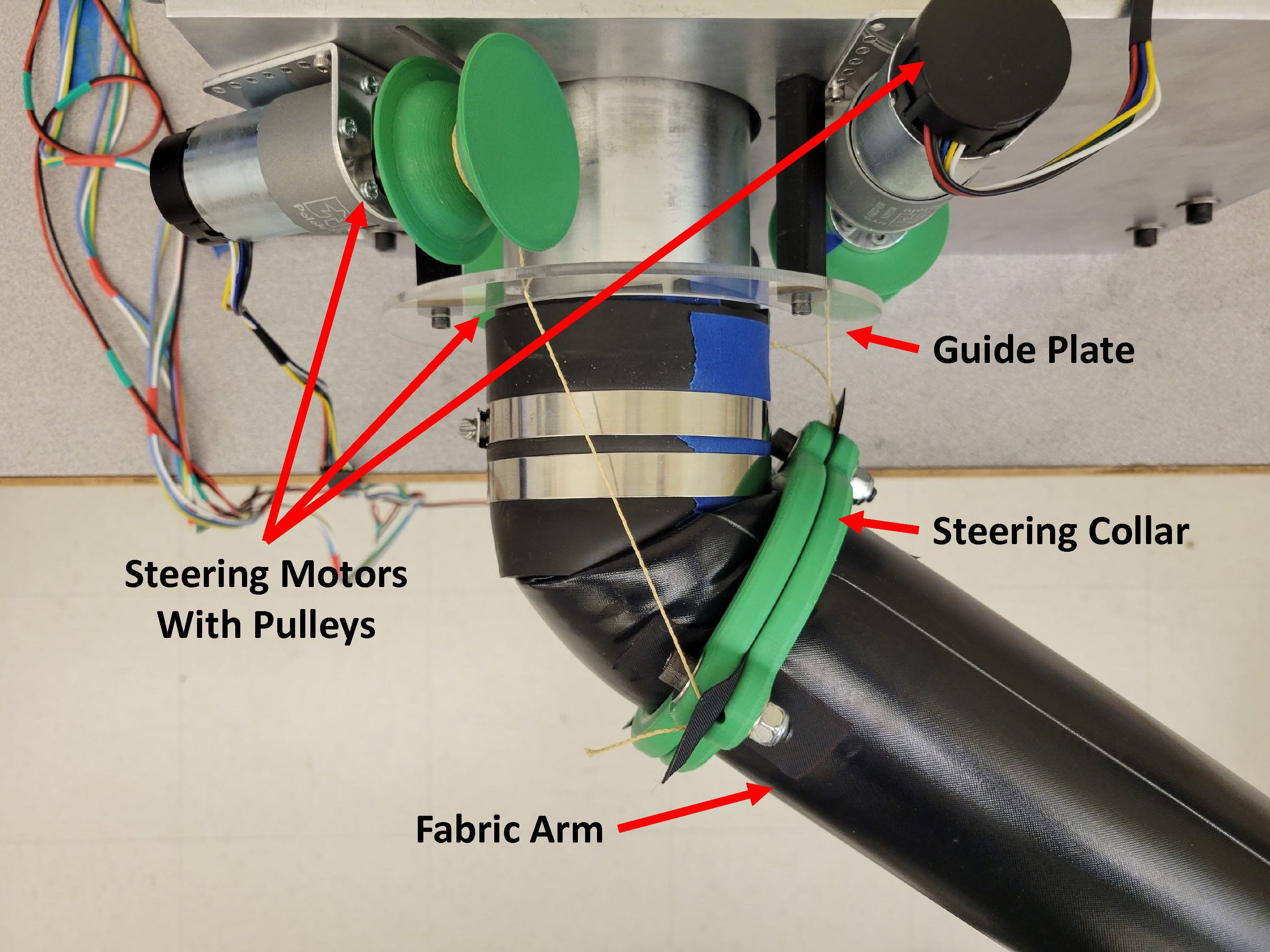}
	\caption{Steering system mounted at the base of the fabric arm can achieve $60^\circ$ of actuation in a single direction.}
	\label{steeringsystem}
	\vspace{-0.1 in}
\end{figure}

\begin{figure} 
	\centering
        \vspace{-0.0 in}
	\includegraphics[width=1\linewidth,]{ 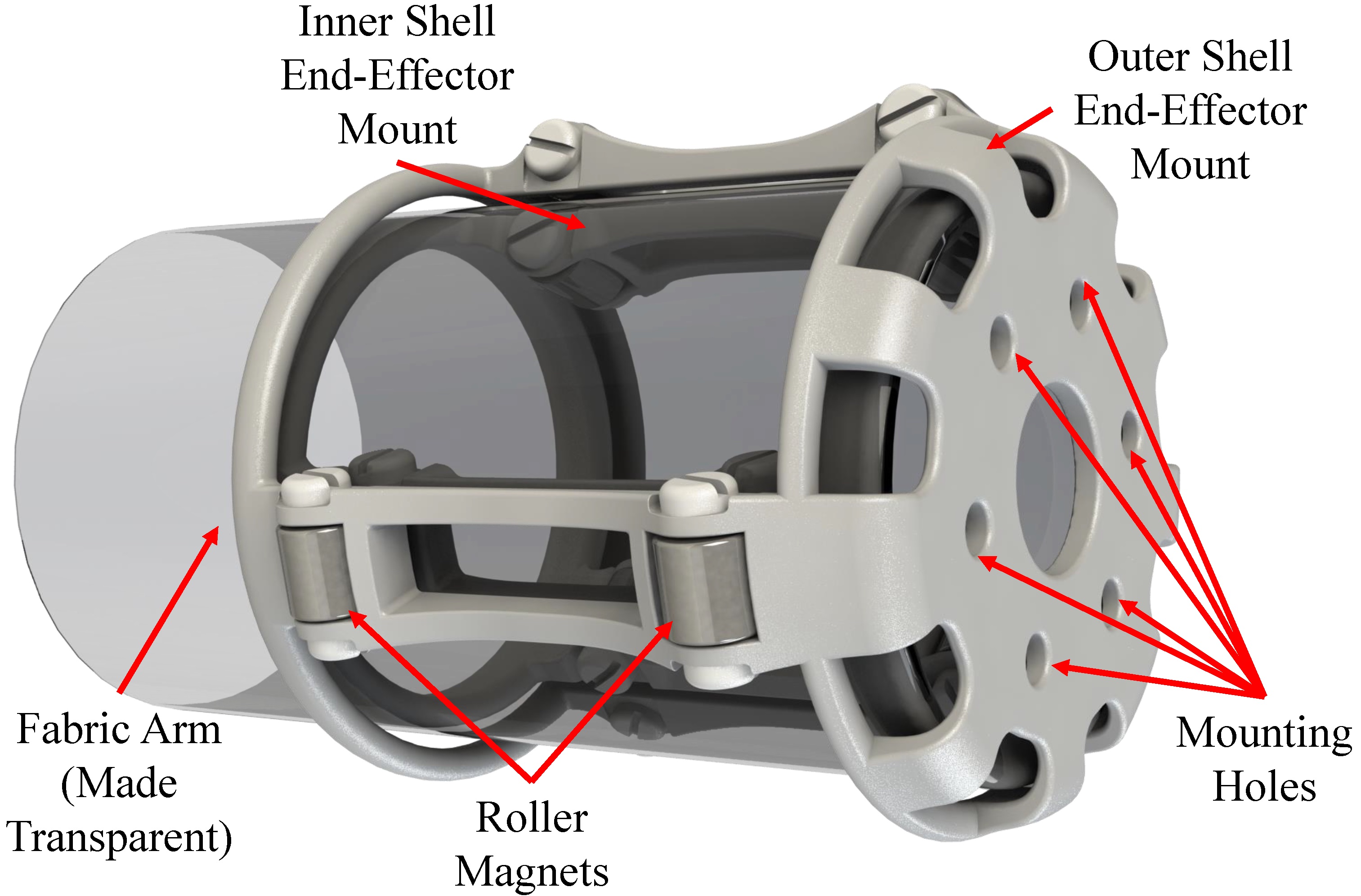}
	\caption{CAD render of the end-effector mount. The mount utilizes an outer mounting shell and an inner shell that interact via roller magnets.}
	\label{endmount}
	\vspace{-0.2 in}
\end{figure}

The soft-growing manipulator arm has four main components, the fabric arm, pressurized enclosure, steering system, and end-effector mount. The system utilizes a similar design to the RoBoa vine robot \cite{der2021roboa}. However, our system was designed to operate at much higher pressures, have a shorter but entirely self-supported arm, and have the steering actuation located at the base of the arm. Our system can reliably operate at pressures below 10~psi (68.9~kPa) relative to atmospheric pressure and can steer up to $60^\circ$ in a single direction relative to the default configuration. These specifications allow our robot to operate in the entirety of a spherical sector workspace with a payload of 1.4~Kg and an arm extension speed of 0.27m/s at 8~psi (55.2~kPa). The system also utilizes a lightweight soft-gripper end-effector that is beyond the scope of this paper. The overall design of the soft-growing manipulator is displayed in Fig.~\ref{cadmodel} with all major components labeled.
\\
\textbf{Fabric arm:} The fabric arm is made out of a 58" 200 denier heat-sealable coated Oxford. The fabric was obtained from Seattle Fabrics under the part number: FHSO. The fabric is black Oxford with a heat-sealable coating made out of white thermoplastic polyurethane. The fabric is cut into a rectangle, heat-welded into a tube-like shape, and then pulled in on itself to create the arm. The central pulley is connected to the end of the arm via a heavy-duty Kevlar rope. The arm is clamped onto the enclosure using two hose clamps. The arm is 81.28~mm in diameter and 1.2~m in length. These parameters are sufficient to reach and pick all apples within a modern apple orchard while on a mobile robot. The current manufacturing method for the fabric arm results in issues with reaching higher pressures and low fatigue life. Thus, future work will aim to address these issues.
\\
\textbf{Pressurized Enclosure:} Our goal for the overall system is picking an apple within 2s and lifting a payload of at least 1.4~Kg. Since the pressure force acting on the end of the arm is the extension mechanism, and the maximum load of a pressurized fabric tube is dependent on its pressure \cite{cavallaro2003mechanics}, higher air pressures are critical in order to achieve these goals. Previous work has demonstrated a maximum pressure of around 3~psi (20.6~kPa) due to the acrylic-made container \cite{9197094}. Therefore, to reach higher pressures, our design utilizes a machined aluminum extrusion airtight enclosure that houses the central pulley assembly and central motor. The enclosure is designed to withstand 20~psi (137.9~kpa), which has been experimentally verified. Using aluminum also makes the design easier to manufacture and reduces weight.
\\
The central motor that controls the robot's length is a Maxon 24~V DC motor, part number: 148867, with a 12:1 gear ratio. With this motor, the arm has a maximum retraction speed of 0.25~m/s at 3~psi (20.6~kPa) and an extension speed of 0.27~m/s at 8psi (55.1~kPa). Since the current mobile platform design involves dropping apples onto conveyor belts or catchers below the arm right after picking, retraction speed is not as crucial as extension speed. 
\\
\textbf{Steering system:} The steering system is composed of three steering motors with pulleys and cables, a guide plate, and a steering collar, as shown in Fig.~\ref{steeringsystem}. Steering is controlled by three 12V DC motors with 150:1 gear ratios. The motors use heavy-duty Kevlar cables that are threaded through the guide plate and connected to the steering collar at the base of the fabric arm to steer the arm. The motors are arranged $120^\circ$ apart on the front mounting plate around the fabric arm. During steering, only two of the steering motors actuate while the third returns to its no-tension configuration. This is because three cables in tension may generate unpredictable behavior. The maximum steering angle in this design is $60^\circ$ in a single direction, but this angle can be increased by increasing the gap between the motors and the steering collar.
\\
\textbf{End-effector mount:} The mount is composed of an inner and outer shell that interact between 
the inside and outside of the arm via roller magnets \cite{luong2019eversion}. The design of the shells and how they connect through the fabric arm are shown in Fig.~\ref{endmount}. The roller magnets provide a strong connection that requires approximately 26.7~N to separate. The mount itself is a lightweight PLA plastic frame (about 275~g, including the roller magnets). The soft robotic gripper (about 450~g) and other components, such as a miniature camera, will be mounted to the front of the outer shell. However, the use and mounting of these components are beyond the scope of this paper.
\subsection{Specification}
\begin{figure} 
	\centering
        \vspace{-0.1 in}
	\includegraphics[width=1\linewidth,]{ 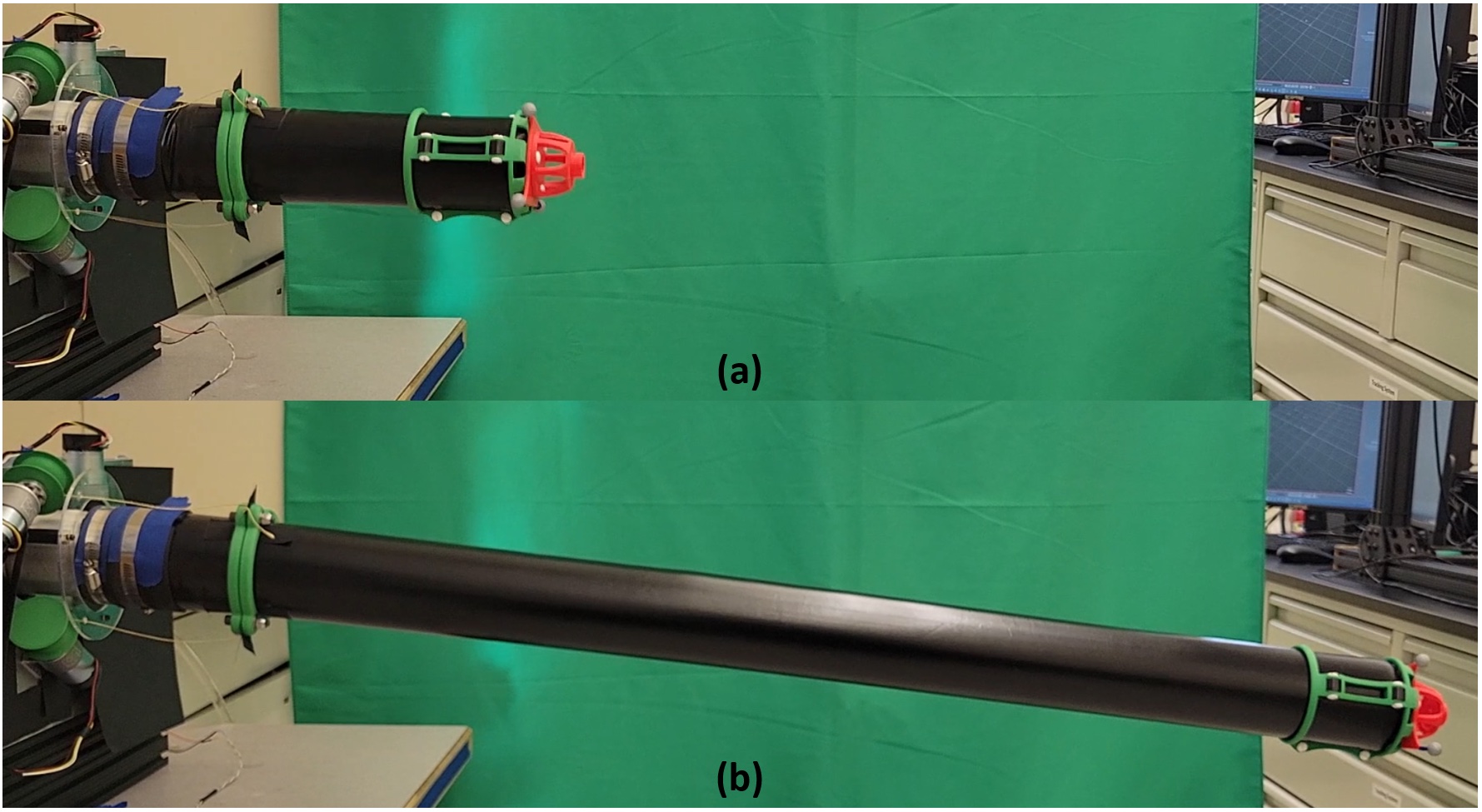}
	\caption{Growing speed test demonstration. (a) Manipulator arm's initial position. (b) Manipulator arm's final position.}
    \vspace{-0.3 in}
	\label{speed}
\end{figure}

Here are the current specifications of the robot, designed to meet the requirements for apple harvesting. Its capabilities can also be improved in further development.
\\
\textbf{Speed:} The growing speed of the arm is determined by the central motor's free running speed, the internal pressure of the system, and the flow rate of the incoming air pressure. Currently, our design has a growing speed of 0.27m/s at 8~psi (55.1~kpa) and a maximum retraction speed of 0.25m/s at 3~psi (20.7~kpa). Snapshots of the arm growing at 8~psi (55.1~kpa) are shown in Fig.~\ref{speed}.
\begin{figure} 
	\centering
	\includegraphics[width=1\linewidth,]{ 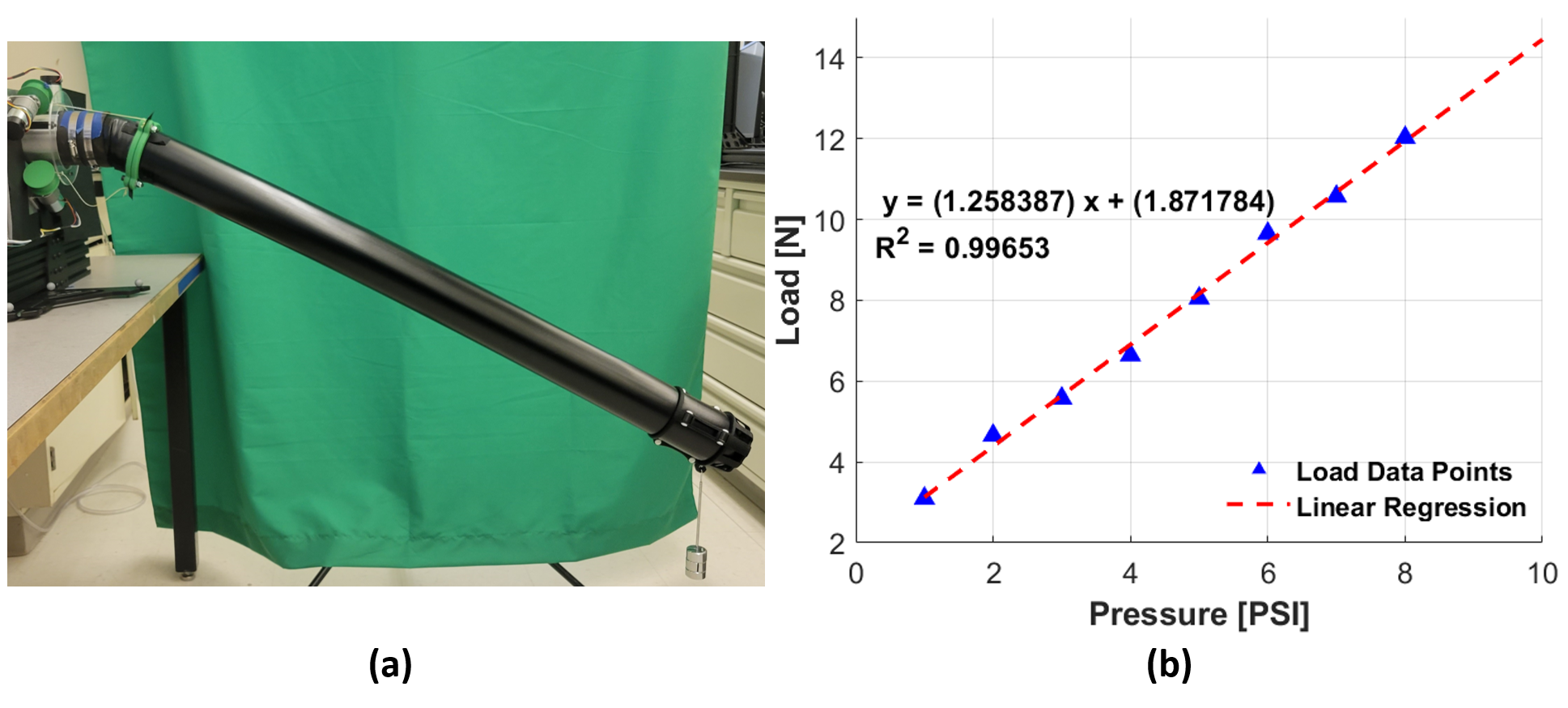}
	\caption{(a) Payload testing where the loaded condition induces buckling at the base of the 1.2~meter long arm. (b) Payload test with pressure from 0 to 8~psi (55.1~kPa). Higher pressures were not tested due to the fabric arm failing at pressures over 8~psi (55.1~kPa).}
    \vspace{-0.1 in}
	\label{payload}
\end{figure}
\\
\textbf{Payload:} The payload of the arm was determined by applying weights to the end of the manipulator arm until it buckled. This process was tested at various pressures while at the full 1.2~m arm length. The buckling point was considered to be the point at which the steering cables could no longer compensate for the arm bending. An example of the buckling point is shown in Fig.~\ref{payload} (a), and the results of the payload test are shown in Fig.~\ref{payload} (b). This process demonstrates that a linear relationship exists between the pressure of the arm and the maximum payload. The current manufacturing method for the fabric arm results in issues with reaching more than 10~psi (68.9~kPa) and low fatigue life. Thus, future work will aim to address these issues.
\\
\textbf{Workspace:} Based on the physical geometry of the system, the manipulator arm has a spherical sector-shaped or cone-like workspace. Specifically, the workspace has a radius of 0.3~m to 1.2~m and a maximum angle of $120^\circ$ centered at the origin.

\section{Static Modeling and Controller Design}
\subsection{Static modeling}
\label{sec:Modeling and control}

\begin{figure} 
	\centering
	\includegraphics[width=1\linewidth,]{ 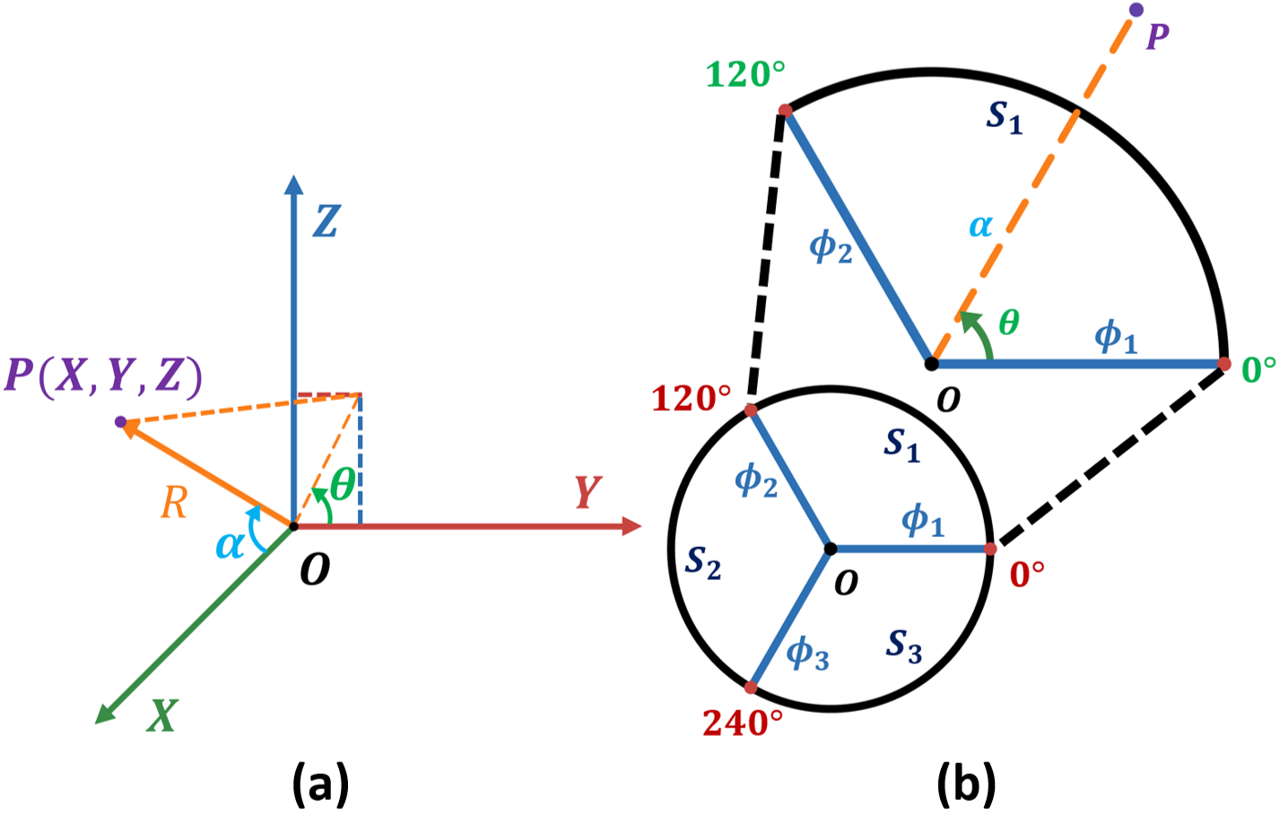}
	\caption{Diagrams of relevant parameters for modeling the soft-manipulator arm. (a) Spherical coordinate parameters. (b) Steering region parameters.}
	\label{modelparameters}
 \vspace{-0.1 in}
\end{figure}

This section introduces a kinematic model that predicts the relationship between the end effector's position and the angular position of three steering motors. Fig.~\ref{modelparameters} (a) shows the parameters used to describe the robotic manipulator's kinematic model in a spherical coordinate system, which is similar to other cable-driven soft continuum robots \cite{walker2013continuous}. The parameters used in the model are as follows:
\\
\textbf{$\phi_{1}, \phi_{2}, \phi_{3}$}: The steering angle of the motors with respect to steering motors 1,2, and 3. In our system, the steering motors are located at $0^\circ,120^\circ,240^\circ$ in the global coordinate system.
\\
\textbf{$S_{1}, S_{2}, S_{3}$}: The steering sections from $0^\circ$ to $120^\circ$, $120^\circ$ to $240^\circ$, and $240^\circ$ to $360^\circ$.
\\
\textbf{$P$}: The position of the end-effector in x, y, and z coordinates.
\\
\textbf{$O$}: Local origin of the arm.
\\
\textbf{$\alpha$}: Arm pitch angle in degrees. $\alpha \in [0^\circ~90^\circ]$
\\
\textbf{$\theta$}: Arm rotation angle in degrees. $\theta \in [0^\circ~360^\circ)$
\\
\textbf{$R$}: Arm length in meters.
\\
\textbf{$k$}: Motor position angle to arm pitch angle coefficient, which can be determined experimentally.
\\
The key difference of this robotic manipulator lies in the following aspects: firstly, it only has a single soft segment, regardless of its length, instead of multiple segments. Secondly, it is guided by only two steering motors instead of three at any given moment. The steering region of the manipulator is divided into three equal sections ($S_{1}, S_{2}, S_{3}$), and each section is controlled by the two steering motors that bound the region, which is shown in Fig.~\ref{modelparameters} (b). The region used is determined by the desired arm rotation angle. While the two engaged motors actuate, the third motor is set to its default configuration so that it does not apply tension to the base of the arm, which may create unpredictable motions.
\\
In order to control the arm to the desired position with the two actuating motors, a relationship between the relative motor steering angles $\phi_{1}, \phi_{2}, \phi_{3}$ and the arm pitch angle $\alpha$ as well as the arm rotation angle $\theta$ was determined. These relationships are shown in eq.~\eqref{e1} and \eqref{e2}. These relationships were formulated by evaluating the point $P$ with respect to an arc defined by the steering region as shown in Fig.~\ref{modelparameters} (b). In this configuration, the arm pitch angle $\alpha$ is the magnitude of the combined $\phi$'s in order for the arm to get to point $P$ from the origin, $O$, while the arm rotation angle $\theta$ is the arc angle. Thus, the arm pitch angle $\alpha$ is the magnitude of the combination of the Y and Z components of the arm pitch angles induced by the two steering motors, and the arm rotation angle $\theta$ is a fraction of the angle range of the section in the Cartesian space.

\begin{equation}
\alpha =
\begin{cases}
k\sqrt{\phi_{1}^{2} -\phi_{1}\cdot\phi_{2} + \phi_{2}^{2}} & \text{if $0^\circ\leq\theta<120^\circ$}\\ 
k\sqrt{\phi_{2}^{2} -\phi_{2}\cdot\phi_{3} + \phi_{3}^{2}} & \text{if $120^\circ\leq\theta<240^\circ$}\\ 
k\sqrt{\phi_{1}^{2} -\phi_{1}\cdot\phi_{3} + \phi_{3}^{2}} & \text{if $240^\circ\leq\theta<360^\circ$}\\
\end{cases}
\label{e1}
\vspace{-0.1 in}
\end{equation}

\begin{equation}
\theta = 
\begin{cases}
120^\circ\phi_{2}/(\phi_{1}+\phi_{2}) & \text{if $0^\circ\leq\theta<120^\circ$}\\
(120^\circ\phi_{2}+240^\circ\phi_{3})/(\phi_{2}+\phi_{3}) & \text{if $120^\circ\leq\theta<240^\circ$}\\
(360^\circ\phi_{1}+240^\circ\phi_{3})/(\phi_{1}+\phi_{3}) & \text{if $240^\circ\leq\theta<360^\circ$}\\
\end{cases}
\label{e2}
\end{equation}

\subsection{Controller design}
\begin{figure*} 
	\centering
    \vspace{0 in}
	\includegraphics[width=1\linewidth,]{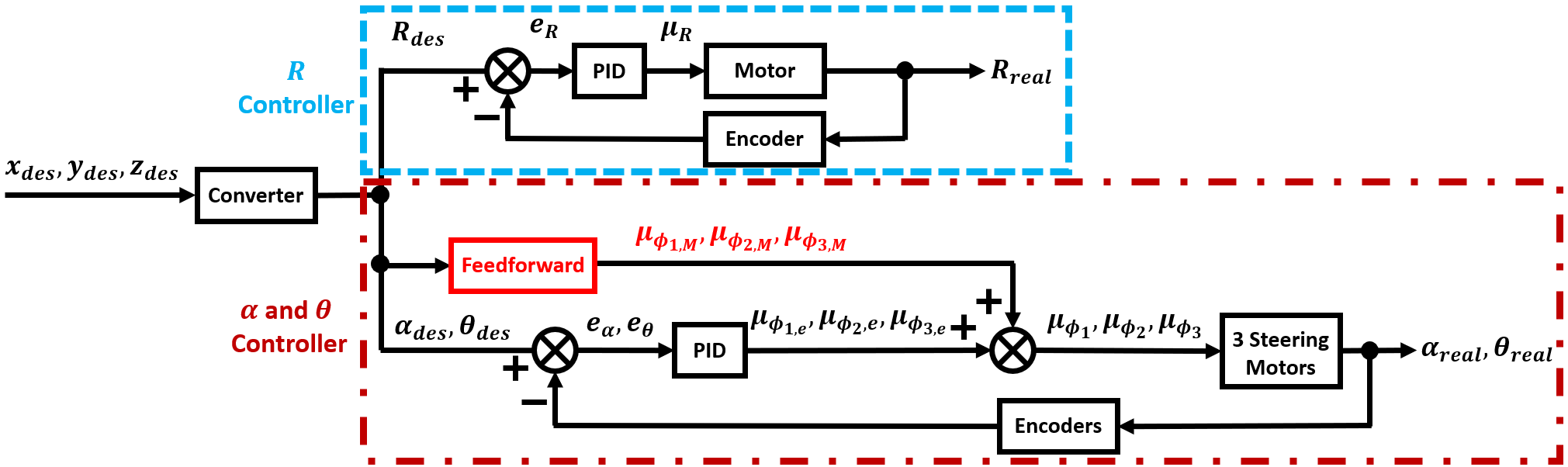}
	\caption{Control loop block diagram for positional control of the soft-growing manipulator arm. The control loop is separated into two separate controllers, one for arm length, $R$, and one for arm pitch angle, $\alpha$, and rotation angle, $\theta$.}
	\label{controllerdiagram}
	\vspace{-0.1 in}
\end{figure*}
The control of the soft-growing manipulator is split into two independent controllers. The first controls the arm pitch angle $\alpha$ and arm rotation angle $\theta$, while the second controls the arm length, shown in Fig.~\ref{controllerdiagram}. The parameters in the diagram are as follows:
\\
\textbf{$x_{des}, y_{des}, z_{des}$}: The desired 3D position of the end-effector in x, y, and z respectively. 
\\
\textbf{$R_{des}, \alpha_{des}, \theta_{des}$}: The desired arm length, pitch angle, and rotation angle respectively based on the desired point.
\\
\textbf{$e_{R}, e_{\alpha}, e_{\theta}$}: Error of the arm length, pitch angle, and rotation angle respectively.
\\
\textbf{$\mu_{R}, \mu_{\phi_{1}}, \mu_{\phi_{2}}, \mu_{\phi_{2}}$}: Updated motor positions for the center motor and the three steering motors respectively. 
\\
\textbf{$\mu_{\phi_{1,e}}, \mu_{\phi_{2,e}}, \mu_{\phi_{2,e}}$}: Updated motor positions for the three steering motors respectively without the feedforward terms. 
\\
\textbf{$\mu_{\phi_{1,M}}, \mu_{\phi_{2,M}}, \mu_{\phi_{2,M}}$}: feedforward motor positions for the three steering motors respectively based on the kinematic model. 
\\
\textbf{$R_{real}, \alpha_{real}, \theta_{real}$}: The actual arm length, pitch angle, and rotation angle respectively.
\\
Both controllers use basic PID closed-loop control. Additionally, the first controller also incorporates feedforward terms, which estimate control inputs using the kinematic model from eq.~\eqref{e1} and \eqref{e2}. This addition of feedforward terms should result in faster convergence rates and less steady-state error compared to the controller without the feedforward terms (Please see Section~\ref{sec:Experiment} C). Here are the equations that demonstrate the control inputs of these controllers: 
\begin{equation}
\begin{aligned}
&\mu_{R} = PID(e_{R}) \\
&\mu_{\phi_{1,2,3}} = PID(e_{\alpha},e_{\theta})+\mu_{\phi_{1,2,3,M}}(\alpha_{des}, \theta_{des})
\label{Controller}
\end{aligned}
\end{equation}
Where $PID$ is the PID controller block for each dc motor, and $\mu_{\phi_{1,2,3,M}}(\alpha_{des}, \theta_{des})$ are the feedforward terms given $\alpha_{des}, \theta_{des}$, generated from eq.~\eqref{e1} and \eqref{e2}.

\section{Experiment and Analysis}
\label{sec:Experiment}
\subsection{Experimental Setup}
\begin{figure} 
	\centering
        \vspace{-0.1 in}
	\includegraphics[width=1\linewidth,]{ 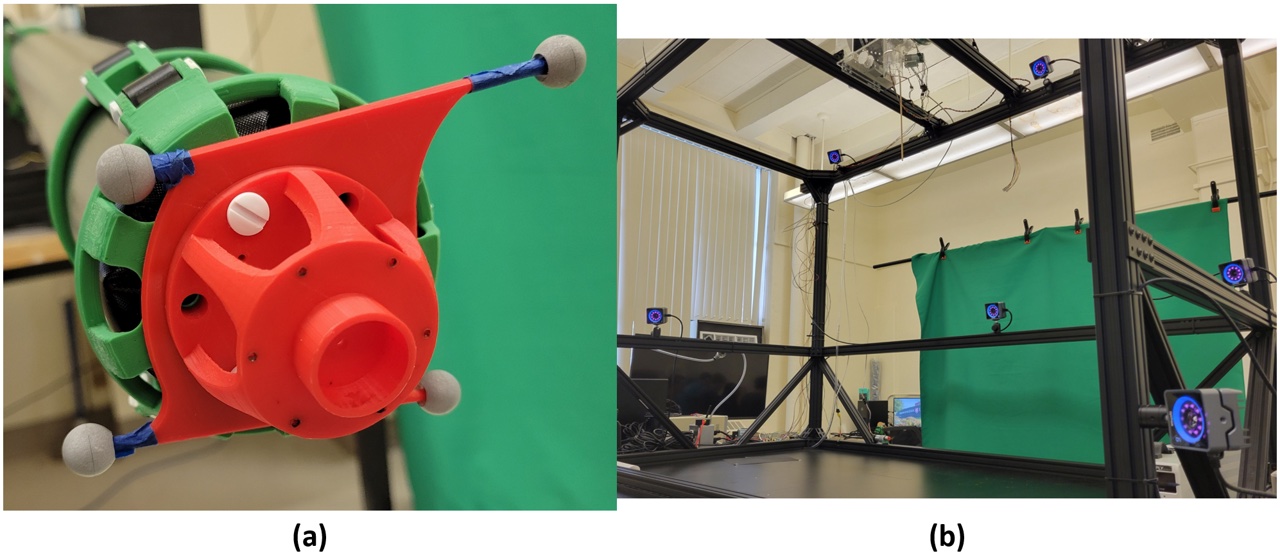}
	\caption{Experimental setup (a) End-effector mount with reflective markers. (b) Motion capture cameras set up around the manipulator arm.}
	\label{testsetup}
	\vspace{-0.2 in}
\end{figure}

To obtain real-time global x, y, and z positional coordinates of the end-effector, an Optitrack motion capture camera system with an internal sync frequency of 120Hz was used. In this setup, six PrimeX 13 cameras were mounted at different locations surrounding the robot, and reflective markers were mounted on the end-effector. The reflective marker end-effector and camera setup are shown in Fig.~\ref{testsetup} (a) and (b) respectively. Positional information was fed to the system via serial port communication between MATLAB and Arduino. MATLAB was used to read the data from Optitrack while all data processing and calculations were done in Arduino. This setup was used for all tests that required the tracking of the arm's movement.

\begin{figure}
	\centering
        \vspace{-0.1 in}
	\includegraphics[width=1\linewidth,]{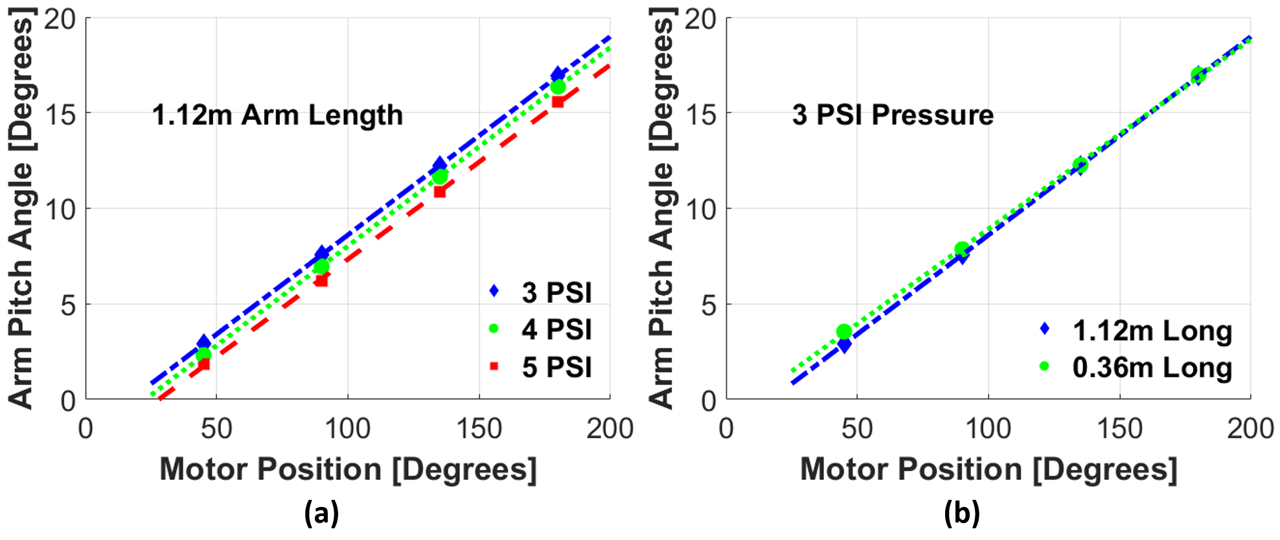}
	\caption{Steering motor angle to arm pitch angle ($\alpha$) for (a) constant arm length and varying pressure and (b) constant pressure and varying arm length.}
	\label{steeringplots}
	\vspace{-0.1 in}
\end{figure}

\begin{figure}
	\centering
        \vspace{-0.0 in}
	\includegraphics[width=1\linewidth,]{ 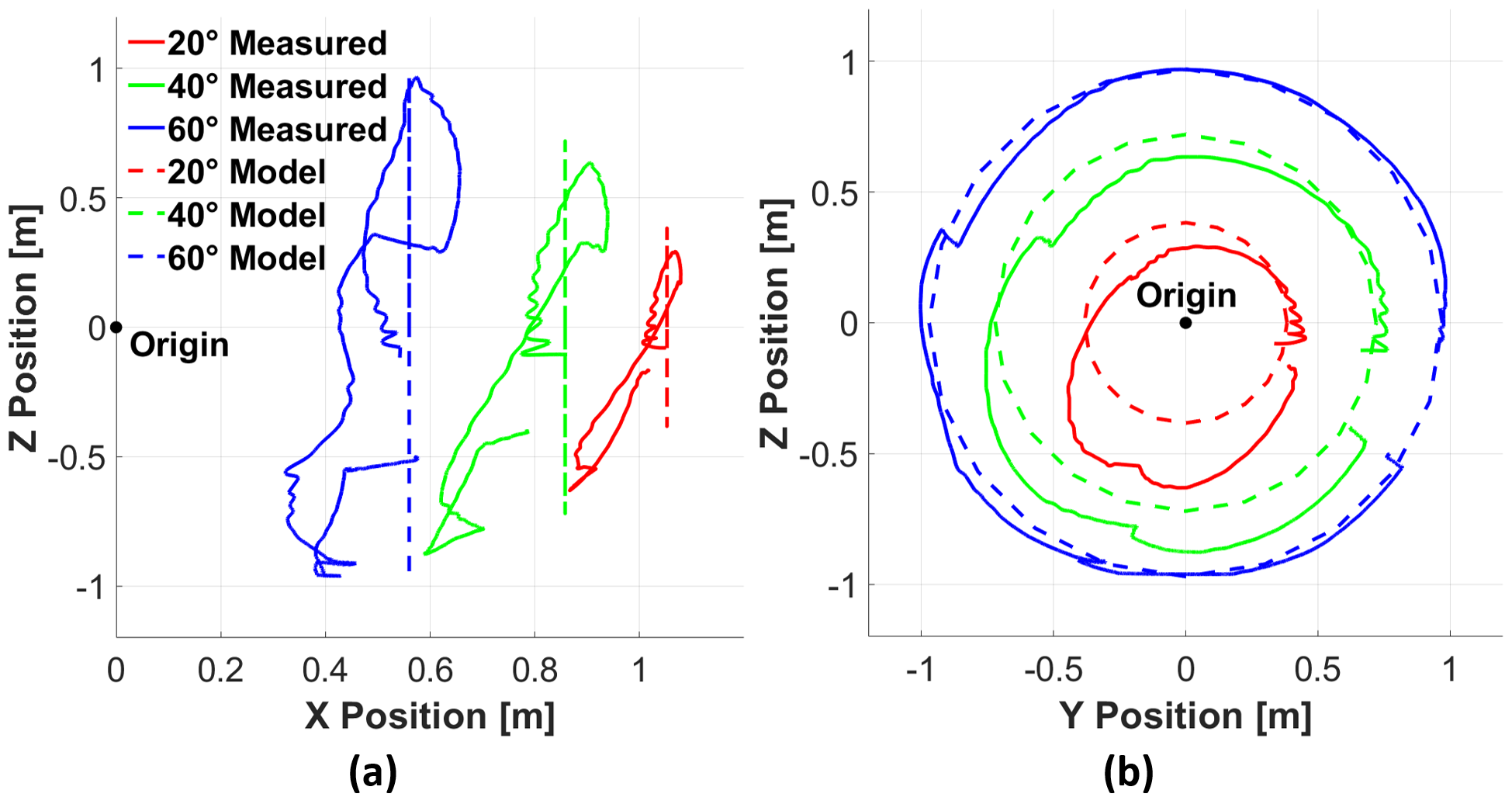}
	\caption{Comparison of the experimental and model predicted end-effector's position with 1.12m arm length and 3~psi (20.7~kpa) arm pressure. (a) Side view. (b) Front view.}
	\label{workspace}
	\vspace{-0.0 in}
\end{figure}
\begin{figure}
	\centering
        \vspace{-0.2 in}
	\includegraphics[width=1\linewidth,]{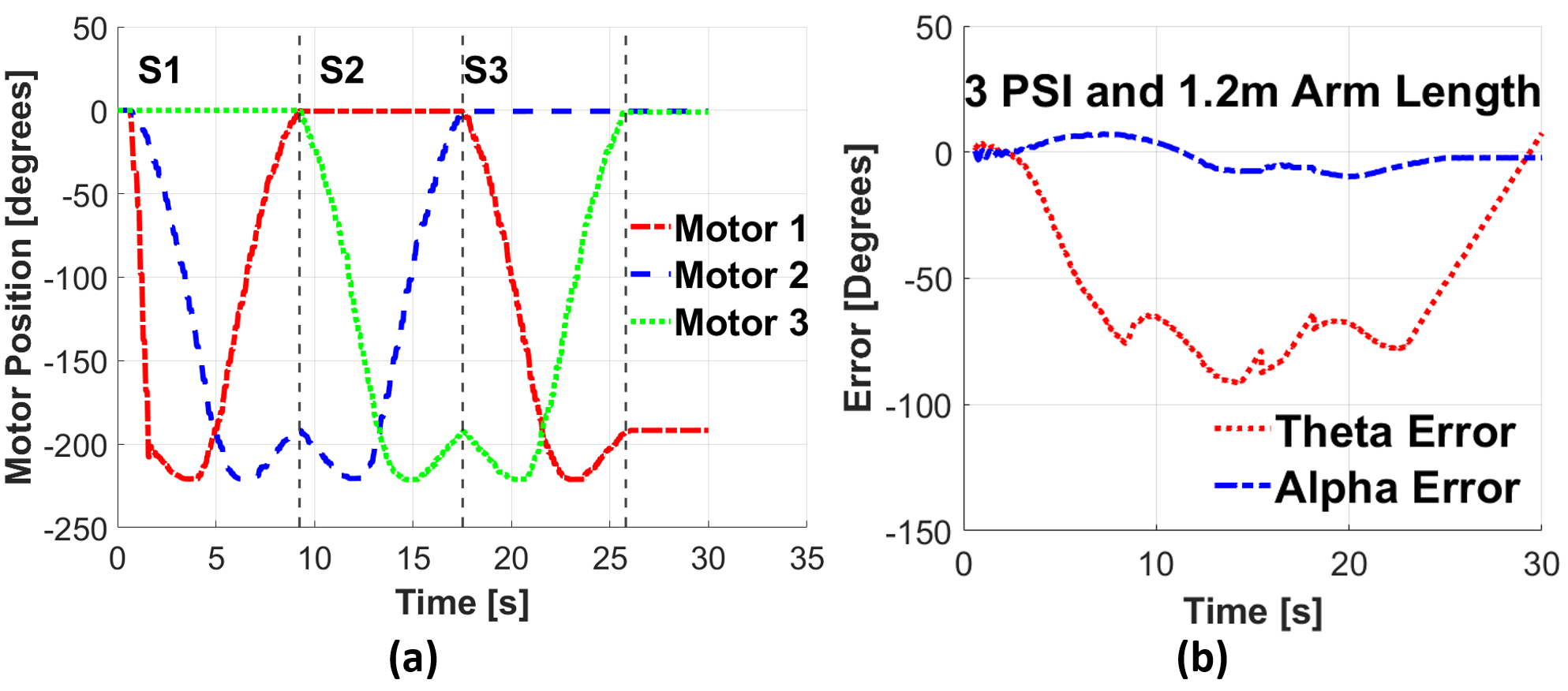}
	\caption{(a) Recorded steering motor angle positions of the manipulator arm while completing a circular motion. (b) The error between the physical system and the model for $\alpha$ and $\theta$ while the arm completes a circular motion.}
	\label{modelcomp}
	\vspace{-0.1 in}
\end{figure}
\begin{figure*}[b]
	\centering
        \vspace{-0.1 in}
	\includegraphics[width=1\linewidth,]{ 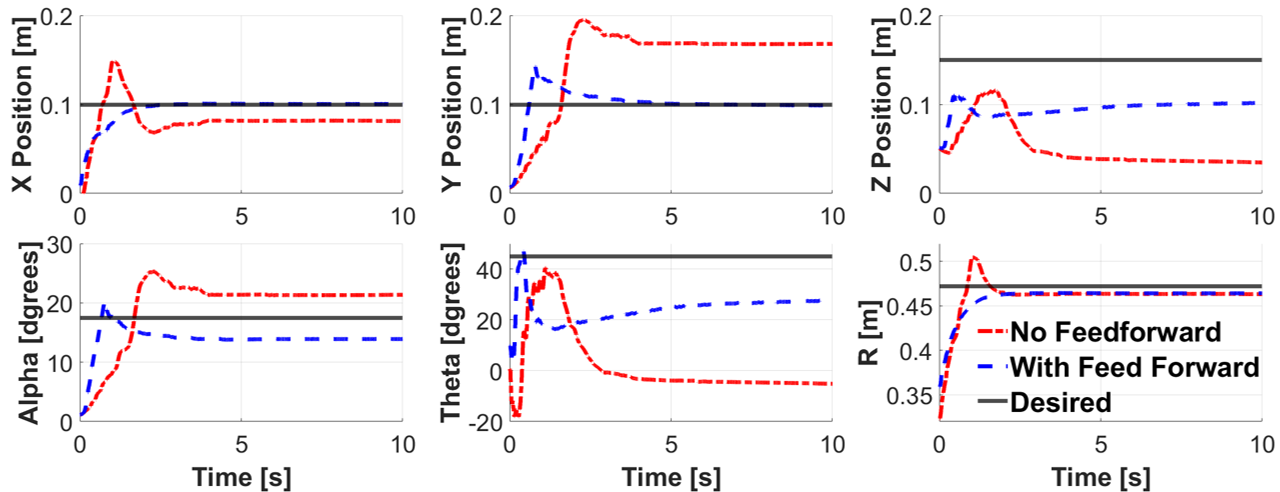}
	\caption{Plots comparing the no feedforward and the with feedforward variations of the positional control loop on the system for the same desired point.}
	\label{positioncontrol}
\end{figure*}

\subsection{Model Verification}
To estimate the kinematics model's parameters, which will be feedforward terms of the feedback loop controller, we conducted two experiments. Firstly, we collected position data as we pulled one steering cable. This data provided the steering motor's angle $\phi_1$ and arm pitch angle $\alpha$ for different arm lengths and pressures. Fig.~\ref{steeringplots} shows that the relationship between the steering motor's angle and arm pitch angle is linear and that $K$ from eq.~\eqref{e1} is approximately 0.104. These results also show that this value is not affected by the input pressure value and length of the robot when the steering system is fixed.

After understanding the arm pitch angle $\alpha$ model, we programmed the arm to do a circular motion with three arm pitch angles and then compared the results. To make the arm move in a circular motion in 3D space, we implemented eq.~\eqref{e1} and \eqref{e2} with each steering motor's position controller. These positions were fed to the system as the actual position of the end-effector was recorded. The results of this process are shown in Fig.~\ref{workspace}. This process displays the manipulator arm's ability to encompass almost the entire range predicted by the model. However, the actual end effector's position cannot reach the area at $\theta$ from $-33^\circ$ to $0^\circ$ degrees. This is due to the arm bending because of gravity. The problem is exacerbated by the configuration of the steering motors, where one of the motors is located at 0 degrees and has a pulling cable that is purely horizontal to the center. Thus, section $S3$ can not contribute any force to compensate for gravity, making the arm unable to reach the upper bounds of this section.

Furthermore, Fig.~\ref{modelcomp} (a) illustrates the three steering motor angles in the circular motion case. As the plot demonstrates, only two steering motors are pulling cables in their corresponding sections $S1, S2, S3$. Fig.~\ref{modelcomp} (b) shows the difference between the arm pitch angle $\alpha$ and the rotation angle $\theta$ of the model prediction and the experimental results, revealing a significant bias in $\theta$ because the model eq.~\eqref{e2} does not account for the shift caused by gravity.

\subsection{Position control}
To verify the benefits of applying the kinematic model to the position controller, we also ran the control loop without the feedforward terms from the default configuration to the same desired point. The feedforward terms can provide an initial three steering motor's angle from the model (eq.~\eqref{Controller},~\eqref{e1},~\eqref{e2} with $k = 0.104$) by giving a desired position. The results of this test are shown in Fig.~\ref{positioncontrol}. During these tests, all PID tuning gains for the DC motors in both controllers were identical. From this process, the controller with the feedforward terms had significantly smaller steady-state errors and setting time compared to the controller with no feedforward terms for both Cartesian and spherical coordinates. While tuning gains can improve the performance of the controller without the feedforward term, such as increasing Ki to reduce steady-state error, these adjustments are unlikely to have as big of an impact as the inclusion of the feedforward terms. It is noticeable that there is an unignorable offset for the arm rotation angle $\theta$ and Z axis with both controllers, which is caused by the bending of the arm due to gravity. The current configuration of three steering motors does not allow for the system to compensate for this bending, resulting in this offset. The $\theta$ plot also demonstrates non-minimum phase behavior for the no feedforward terms controller, because $\theta$ is defined as $0^\circ$ and $360^\circ$ at the horizontal axis. Thus, when the arm begins to follow the circular path, the error is not large enough to significantly actuate motor 2. This results in the arm momentarily dipping into section $S3$ due to gravity. Thereby the $\theta$ value decreases until the error is large enough for motor 2 to generate a large enough torque to lift the arm.

\section{Conclusion} 
\label{sec:conclusion}
In this paper, we demonstrate the design, modeling, and control of a soft-growing manipulator arm that we believe will be useful for future orchard operations. We describe the system's design and its relevant specifications for apple harvesting operations in modern orchards. Specifically, the system has the ability to grow linearly up to 1.2 meters at a maximum speed of 0.27 m/s at 8~psi (55.1~kpa), a decoupled linear and steering controller, and a payload of around 1.4~Kg at 10~psi (68.9~kpa). These specifications can be further improved by modifying the design to address limitations and issues. A kinematic model describing the relationship between the steering motors and the steering angle of the arm was developed and experimentally verified. Also, this verification highlighted a design issue that results in the arm being unable to compensate for the bending due to gravity in certain sections. In addition, with the feedforward terms from the kinematic model, the position controller displayed significantly lower steady-state errors and settling time compared to the controller without the feedforward terms. Our ongoing work is further improving the design of the system and controller for more robust applications in apple orchards. For example, we are modifying the steering motor position to avoid the steady-state error induced by bending due to gravity. We will work on creating an optimal controller to reach any point in the workspace faster and with less vibration. Also, we will integrate the system with our existing onboard vision systems for apple detection to obtain both robot and apple's position in the same coordinate system \cite{joseph2016proof}. Lastly, we will integrate the soft-gripper end-effector to complete apple harvesting operations using a mobile robotic platform.

\bibliographystyle{IEEEtran}
\bibliography{Bibliography.bib}
\end{document}